\pdfoutput=1
\documentclass[11pt]{article}

\usepackage{emnlp2021}

\usepackage{amsmath}

\usepackage{times}
\usepackage{latexsym}

\usepackage[T1]{fontenc}
\usepackage[utf8]{inputenc}

\usepackage{microtype}

\usepackage{graphicx}

\usepackage{subcaption}

\newcommand{\pgn}{PGN }
\newcommand{\lfbert}{LF2BERT }
\newcommand{\B}{\fontseries{b}\selectfont}
\usepackage{makecell}

\title{Abstractive summarization of hospitalisation histories with transformer networks}
\author{Alexander Yalunin \\
    Sber AI Lab \\
 \texttt{ale.yalunin@gmail.com} \\\And
 Dmitriy Umerenkov \\
 Sber AI Lab \\
 \texttt{} \\
  \texttt{Umerenkov.D.E@sberbank.ru} \\\And
 Vladimir Kokh \\
 Sber AI Lab \\
 \texttt{kokh.v.n@sberbank.ru}
 }

\begin{document}
\maketitle
\begin{abstract}
In this paper we present a novel approach to abstractive summarization of patient hospitalisation histories. We applied an encoder-decoder framework with Longformer neural network as an encoder and BERT as a decoder. Our experiments show improved quality on some summarization tasks compared with pointer-generator networks.
We also conducted a study with experienced physicians evaluating the results of our model in comparison with PGN baseline and human-generated abstracts, which showed the effectiveness of our model.


\end{abstract}

\section{Introduction}

Each year hundreds of millions of patients are admitted to hospitals worldwide. At the end of hospitalisation the doctors need to make a summary of the hospitalisation notes. This task can be automated by using automated systems that require strictly structuring the hospitalisation notes. At institutions where such strict structure is not imposed on clinical records, this time consuming work is performed manually and often passed to junior medical staff.

\begin{figure}[t]
\begin{center}
\includegraphics[width=\linewidth]{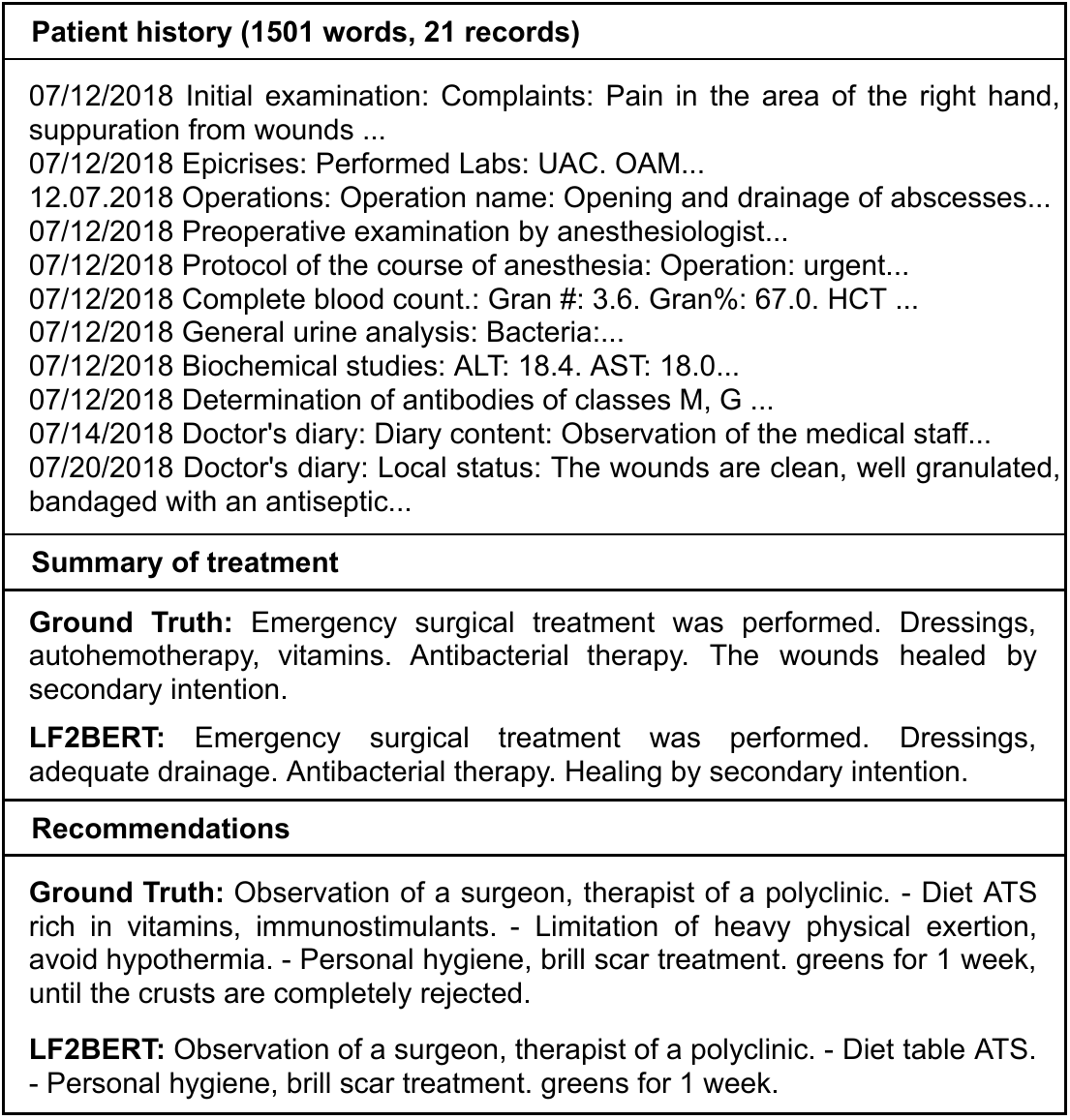}
\caption{Example of patient history with the beginning of records, the corresponding summary of treatment and further recommendations sections.} 
%  The complete example on the original language can be found in Appendix (Figure \ref{figure:case1}).
% \caption{} 
\label{figure:preview}
\end{center}
\end{figure}

Previous works in the medical domain actively use pointer-generator networks for summarization tasks, including summarization of radiology findings, medical dialogues and SOAP notes. In this work we propose LF2BERT - a novel method of neural abstractive medical summarization with an end-to-end transformer model. We use an encoder-decoder framework with a Longformer model as an encoder and BERT as a decoder. As shown in Figure \ref{figure:preview}, given the patient history we try to generate three sections of the discharge note: summary of patient treatment, summary of performed medical examinations, and further recommendations.

We conducted experiments comparing the results of our model with a state-of-the-art deep neural network approach in medical summarization (pointer-generator network) both on automated metrics (ROUGE-1, ROUGE-2, ROUGE-L) and using an independent doctor evaluation.

Our experiments confirm the effectiveness of our approach on automated metrics by considerably improving upon the state-of-the-art neural network baseline. The doctor evaluation showed that the summaries generated by our model are better rated that human-generated summaries for some criterions and sections. 

Summarizing, we make the following contributions in this paper:

• We propose a novel end-to-end transformer-based approach in abstractive summarization of long medical texts.

• We conduct experiments to demonstrate that our transformer-based approach outperforms the state-of-the-art approach in medical summarization – pointer-generator network.

• We conduct doctor evaluation to show that our approach has signiﬁcant clinical validity.

%The rest of the paper is organized as follows. In section 2 we discuss related work and in Section 3 we describe the dataset used for this work. In section 4 we describe in detail our method and baselines. In Section 5 and Section 6 we describe our experiments and discuss the results. Section 7 is dedicated to conclusion.

\section{Related work}

\subsection{Neural abstractive summarization}

Abstractive Text Summarization is the task of generating a short and concise summary that captures the salient ideas of the source text. 

Modern transformer-based models like Pegasus \cite{zhang2020pegasus}, BART \cite{lewis2019bart}, ProphetNet \cite{qi2020prophetnet} have shown significant development in abstractive document summarization on datasets of small length articles like CNN/DailyMail and Gigaword. However, on large document datasets like Pubmed or arXiv they perform less well, partly because their architectures do not process entire documents at once but either truncate them to 512 tokens or process them in fixed-length chunks. This is due to the computational complexity of self-attention increasing quadratically with input sequence length.

There are several modifications of self-attentive models that deal with long documents in NLP. One of them, Longformer \cite{beltagy2020longformer}, uses sliding attention windows and global attention to deal with square complexity. Although it is a powerful model, it cannot be used for summarization alone because it is an encoder. BigBird \cite{zaheer2020big} is an encoder-decoder model that adds random attention to the attention used in Longformer and has shown state-of-the-art results on long-document abstractive summarization on datasets such as arXiv and Pubmed.

In this paper we combined different types of transformer-based models. \cite{rothe2020leveraging} justifies the possibility of combining different pretrained models in the encoder-decoder architecture, as well as that sharing parameters between encoder and decoder results in the best strategy.

In this paper we used an approach proposed by T5 \cite{raffel2019exploring} text-to-text framework where one model is used for multiple NLP tasks. To distinguish what task the model will solve at a given time, a task-speciﬁc prefix is added to the original input sequence. For example, a model trained to translate text from English into multiple other languages could be fed the following sequence to instruct a translation into German: “translate English to German: That is good.”

\subsection{Medical summarization}

Text summarization is actively explored for the medical domain \cite{afantenos2005summarization}, including for summarizing radiology findings, medical dialogues and SOAP notes.

Pointer-generator networks \cite{see2017get} are actively used as the summarization model in the medical domain. They use attention and pointer mechanisms, which allow them to either generate a token from a fixed vocabulary or to copy a token from the source text.

\cite{zhang2018learning} generate impressions by summarizing the study background information and the radiology findings. They use two pointer-generator models for each section.
They show that their abstractive model significantly outperforms the extractive one.
They also conduct an evaluation with a board-certified radiologist and show that 67\% of sampled system summaries are at least as good as the corresponding human-written summaries. 

\cite{liu2019topic} split medical conversations into different topic segments using a rule-based lexical algorithm and train a pointer-generator network augmented with a topic-level attention mechanism. 

\cite{joshi2020dr} summarize medical conversations with an improved pointer-generator network with a penalty that encourages copying rather than generating. On the conducted doctor evaluation their model captures most or all of the information in 80\% of the conversations. 

\cite{krishna2020generating} generate SOAP notes from medical conversations. To be able to summarize long conversations, they use extractive summarization to select noteworthy utterances for different sections, then cluster related extracted utterances and generate an abstractive one-line summary for each cluster. They show that a pretrained BERT is the best encoder for sentences in the extraction phase.

Our work represents the ﬁrst attempt in abstractive summarization of long medical texts with an end-to-end transformer-based model.

\section{Dataset}

\begin{figure*}
    \centering
    \subfloat[Number of words in patient history.\label{fig:1a}]
        {\includegraphics[width=0.33\textwidth]{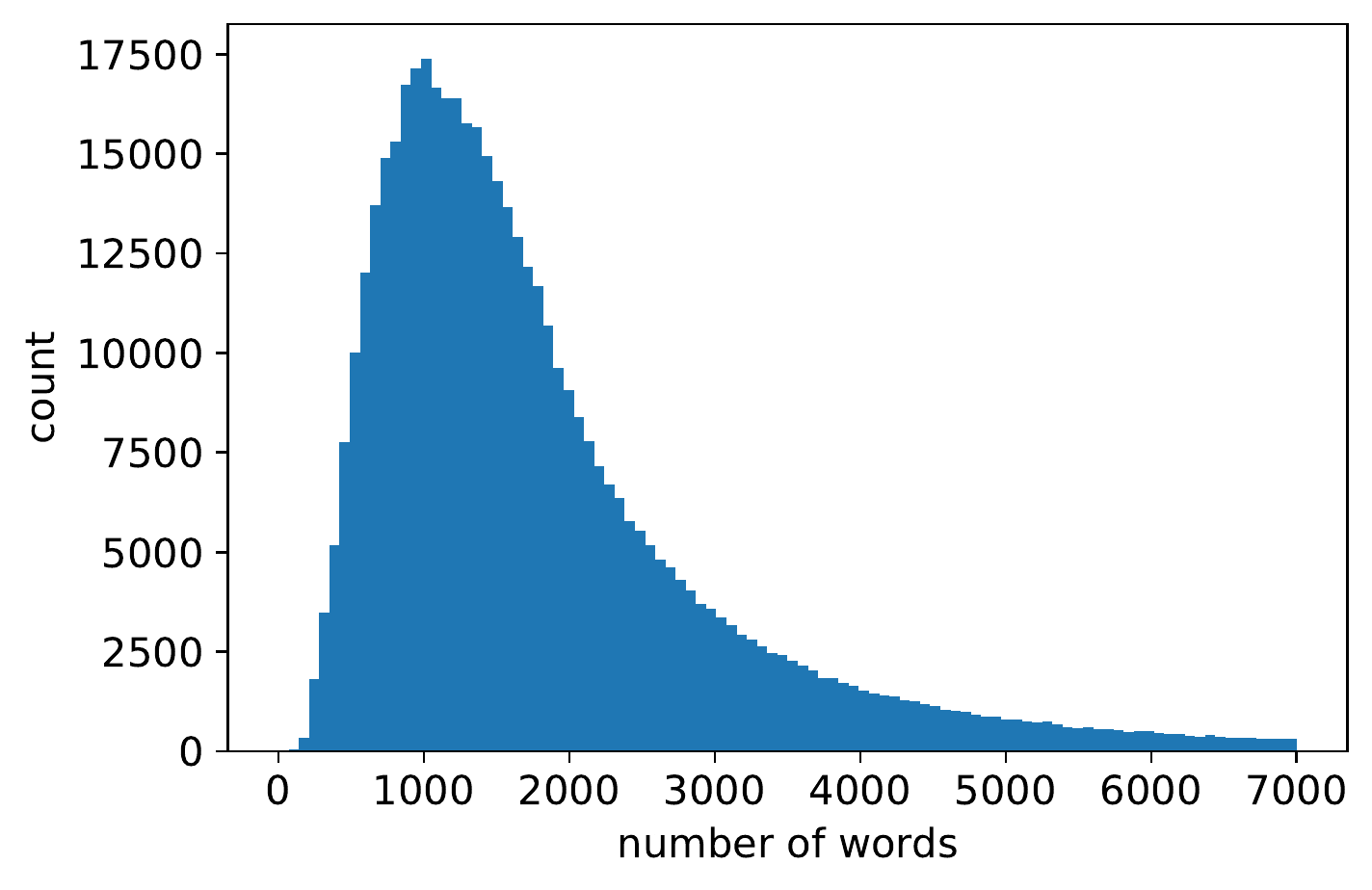}}\hfill
    \subfloat[Number of words in Treatment section.\label{fig:1b}]
        {\includegraphics[width=0.33\textwidth]{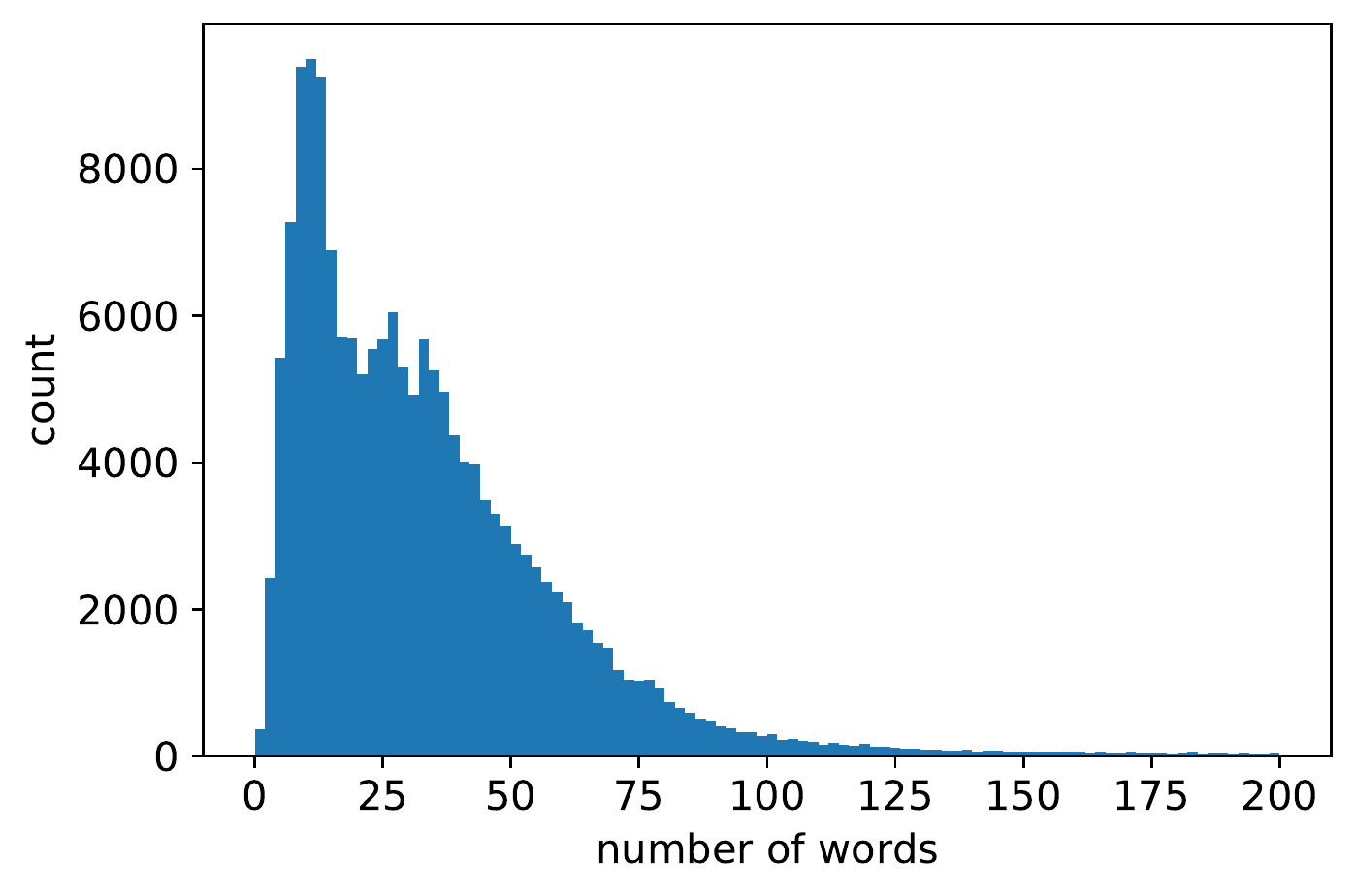}}\hfill
    \subfloat[Number of words in Recommendations section.\label{fig:1c}]
        {\includegraphics[width=0.33\textwidth]{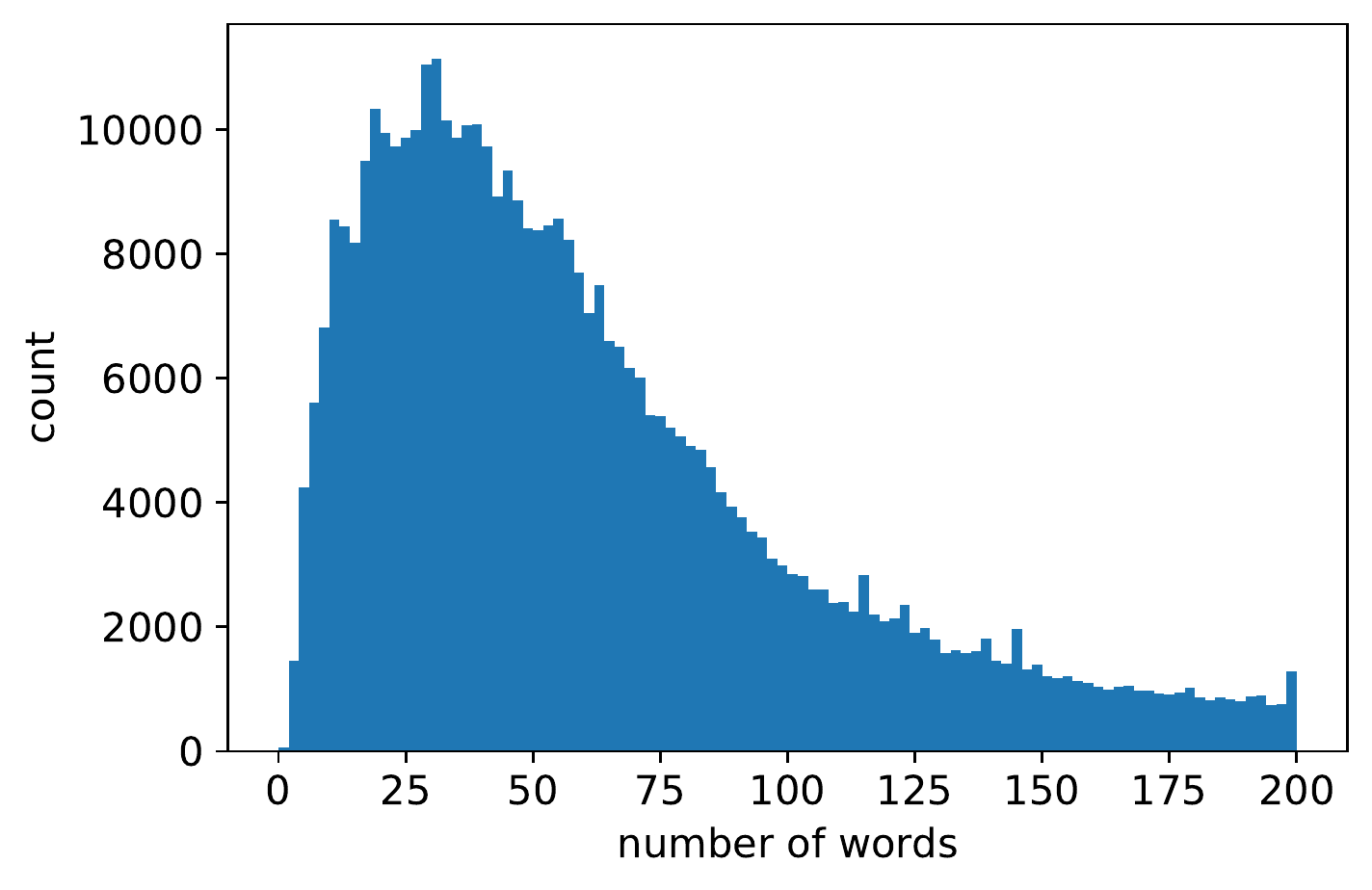}}
    \caption{Distribution of number of words in patient history and in two sections of the discharge note.} \label{fig:worddistributions}
\end{figure*}

We collected a dataset of 561,000 anonymized patient histories from 300,000 unique patients resulting in over 8,912,000 singular records. These records contain information about the patient treatments, medical examinations, surgeries, symptoms, objective status, etc. A patient history on average contains 16 records. The average number of words in patient history is 2080 (Figure \ref{fig:worddistributions}).
The data was collected from dozens of hospitals from a single Russian city from 2014 to 2020.

%Before training model to generate section of discharge notes we use nearly 9 million of raw unlabeled records for pretraining on MLM task.  The average number of words is 135.
Each patient history ends with a discharge note that contains multiple sections including final diagnosis, anamnesis, objective status, etc. 
%several fields about patient like age, height and weight. 
For the purpose of this work we decided to generate the following sections:
\begin{itemize}
    \setlength\itemsep{-0.3em}
    \item Summary of patient treatment (Treatment).
    \item Summary of performed medical examinations (Performed Labs).
    \item Further recommendations (Recommendations).
\end{itemize}

In some summaries of patient treatment and medical examinations the information that appeared in the discharge note was not present in the patient history. We explain this by the fact that our data was incomplete, i.e. not all of the records were entered into the medical system and some discharge notes were based on handwritten notes. 

We decided to use only cases where the discharge summary actually summarizes the available history and does not contain new information. To select such cases we introduce a score that calculates the fraction of medical concepts in the summary that appear in the original texts. 

As the source of medical concepts we used the Unified Medical Language System (UMLS). Specifically a subset of UMLS, Medical Dictionary for Regulatory Activities (MedDRA) was used to obtain a list of medical terms in Russian.

To calculate the above-mentioned score we first lemmatized both the patient history and the summary, then counted the medical concepts that appear in them. Finally we calculated the ratio of the number of medical concepts in the interaction over the number of medical concepts in the summary. We choose 0.5 as the cut-off threshold for including a particular history in our dataset. 

The statistics of our final dataset are shown in Table \ref{tab:Dataset_statitstics}. 

%There are significant number of discharge notes without summary of patient history or medical examinations, but still contain a recommendations section.
%After applying this filtering we got 175,000, 469,000 and 105,000 samples for 1) the summary of patient history, 2) the summary of medical examinations and 3) the further recommendations respectively. 

\begin{table}[t]
\centering
\small
{\begin{tabular}{l|c|c}
& \thead{Number of histories \\ containing this section}  & \thead{Average \\ word count} \\
\hline 
Treatment & 175.000 & 35\\
Performed Labs & 105.000 & 195 \\
Recommendations & 469.000 & 74 \\
\hline 
\end{tabular}}
\caption{Dataset statistics} 
\label{tab:Dataset_statitstics}
\end{table}

%The average number of words in patient history is 2080. The average number in the target sequences are 35, 74 and 195 words in patient history, medical examinations and recommendations.

We split our datasets by patients into train and validation using 97 percent of the data for training and 3 percent for validation. 
%resulting in 727,000 and 22,400 samples in train and val sets respectively.

Additionally we used two proprietary Russian language anonymized EHR datasets. The first contains information about 2,248,359 visits of 429,478 patients to two networks of private clinics from 2005 to 2019 and does not contain hospital records. The second dataset comes from a regional network of public clinics and hospitals and contains 1,728,259 visits from 2014 to 2019 of 694,063 patients.
These additional datasets were used for pretraining our model on masked language modelling task.

\section{Method}

We start with describing the formal definition of the task. Given the patient history represented as sequence of tokens $x=\left\{{x}_{1}, \ldots, {x}_{n}\right\}$ of length $n$, we try to generate the text of one of the three sections of discharge note represented as a sequence of tokens $y=\left\{{y}_{1}, \ldots, {y}_{m}\right\}$ of length $m$. 
 
Given a patient history $x$ and the text of target section $y$, the task is to learn a sequence-to-sequence model with parameters $\theta$ that maximizes the negative conditional log-likelihood:

\[\boldsymbol{\theta}^{*}=\arg\max_{\boldsymbol{\theta}}\sum_{(\boldsymbol{x},\boldsymbol{y})}\log p(\boldsymbol{y}\mid\boldsymbol{x};\boldsymbol{\theta})\]

%\[\boldsymbol{\theta}^{*}=\arg \max _{\boldsymbol{\theta}} \sum_{(\boldsymbol{x}, \boldsymbol{y})} \log p(\boldsymbol{y} \mid \boldsymbol{x} ; \boldsymbol{\theta})\]

\subsection{Pointer-Generator Network}

Previous works actively use pointer-generator networks for the task of summarizing radiology findings, medical dialogues and SOAP notes. A pointer-generator network is an encoder-decoder model with attention and pointer mechanisms. A Bi-directional Long Short-Term Memory \cite{afantenos2005summarization} (Bi-LSTM) network is used in the encoder part and a separate unidirectional LSTM is used in the decoder. The encoder accepts the sequence of tokens and produces a hidden state. Given this hidden state the decoder generates the tokens of the target sequence step-by-step. Additionally at each step an attention distribution is calculated across the input words. This distribution is used to calculate a context vector by taking a weighted average of the output embeddings of the encoder. At each step the model calculates a switching probability to either generate a token from a fixed vocabulary or to copy a token from the source text. This particular feature allows the model to handle out-of-vocabulary words, which are especially frequent in medical domain.

\subsection{LF2BERT}
While the pointer-generator network demonstrates strong performance in medical summarization tasks, due to the multihead self-attention mechanism, resent transformer-based models achieve state-of-the-art results for the summarization task in the general NLP domain. This is due to the increased size of the model and the power of long pretraining on language modelling tasks.  

%We start with explaining the encoder and decoder part of our LF2BERT model.

The base model that we used in our approach is BERT \cite{devlin2018bert}. BERT uses masked language modeling for pretraining bidirectional word representations and provides contextualized word representations during the fine-tuning stage.
However processing long sequences such as hospitalisation histories with transformer-based models like BERT is very computationally expensive due to quadratic complexity of the self-attention mechanism with regard to input sequence length.
There are several approaches to handle large input sequences and the quadratic complexity attention problem \cite{kitaev2020reformer, choromanski2020rethinking, beltagy2020longformer}. 

In this work we used the Longformer model \cite{beltagy2020longformer}, with an attention pattern which employs a ﬁxed-size window attention surrounding each token. In addition, a fixed number of tokens attend to all tokens across the sequence to calculate the so-called “global” attention. 
We use Longformer as an encoder which accepts long sequences of tokens of patients histories and a pretrained BERT as a decoder to handle short sequences of tokens of target section.

\cite{afantenos2005summarization} show the efficacy of initializing the encoder and decoder in transformer-based sequence-to-sequence model from pretrained checkpoints. We used this idea for our LF2BERT model pretraining encoder and decoder separately on a corpus of medical texts.

\subsection{Pretraining BERT}
We start with pretraining BERT. First we used our pretraining datasets to create a custom cased tokenizer with vocabulary size of 40.000 using Byte-Pair Encoding technique \cite{sennrich2015neural}.
%to mitigate the out-of-vocabulary issue.
We randomly initialized a BERT-Base model with our vocabulary size, with default parameters from \cite{devlin2018bert} (12 layers, 12 attention heads, hidden size=768, input length=512). We pretrained the BERT model on the pretraining datasets using a basic masked language modeling task with a 0.15 probability of each token to be masked. By using a tokenizer fitted on medical texts and medical pretraining datasets we expect our model to learn domain-specific token embeddings. 
We pretrained BERT for 3 epochs with mixed precision, starting learning rate of 5e-5, linear learning rate decay, Adam optimizer \cite{kingma2014adam} and a batch size equal to 192. The whole process took 4 days on 16 NVIDIA Tesla V100 GPUs. This model is used as an initial checkpoint for the decoder in LF2BERT model. 
We used the same model to initialize the Longformer model before additional pretraining.

\subsection{Pretraining Longformer}

We used our pretrained BERT to initialize the Longformer. All the model weights were copied without any change, except for positional embeddings. Similarly to \cite{beltagy2020longformer} we initialize 8192 positional embeddings by copying 512 positional embeddings from the pretrained BERT multiple times. 
After this initialization the model already has quality token embeddings but lacks positional embeddings for positions greater than 512, thus additional pretraining on larger sequences is required.
We additionally pretrained Longformer on MLM task using our pretraining datasets. The mask probability was the same as our BERT pretraining had. According to \cite{beltagy2020longformer} the model needs to learn the local context ﬁrst before learning to utilize the longer context. Thus before training we sorted our dataset by the number of tokens, so that the model first gets familiar with short patient histories and then adapts to longer histories. This feature significantly reduces the computational cost and therefore pretraining time. We pretrained it for 3 epochs and a batch size equal to 4. The whole process took 3 days on 1 NVIDIA Tesla V100 GPU. Additionally we used gradient checkpointing \cite{chen2016training} to reduce memory usage.

\subsection{Training LF2BERT}
We used our pretrained Longformer and BERT to initialize LF2BERT model. We tied encoder and decoder word embeddings by making a reference from decoder input embeddings to the encoder ones.
The maximum input and output lengths were set to 8192 and 256 tokens respectively.
We finetuned the model to predict a sequence of tokens of target section given the sequence of tokens of patient history by optimizing the standard teacher forced cross entropy loss. 
We trained the LF2BERT for 3 epochs with mixed precision, gradient checkpointing, starting learning rate of 5e-5, linear learning rate decay, Adam optimizer and batch size equal to 6. The whole process took 2 days on 8 Tesla V100 GPUs.

\begin{table*}[ht]
\centering
{\begin{tabular}{llccc}

\hline 
Section & Model & ROUGE-1 F1 & ROUGE-2 F1 & ROUGE-L F1  \\
\hline 
Treatment & Random & 0.053 & 0.012 & 0.044 \\
& \pgn & 0.563 & 0.460 & 0.542 \\
& \lfbert & \B 0.670 & \B 0.564 & \B 0.645 \\
\hline 
Performed Labs & Random & 0.123 & 0.026 & 0.078 \\
& \pgn & 0.266 & 0.185 & 0.227 \\
& \lfbert & \B 0.484 & \B 0.386 & \B 0.436 \\
\hline 
Recommendations & Random & 0.164 & 0.037 & 0.117 \\
& \pgn & 0.436 & 0.320 & 0.394 \\
& \lfbert & \B 0.644 & \B 0.552 & \B 0.609 \\
\hline 

\end{tabular}}
\caption{ROUGE scores for different sections and models.}
\label{tab:valid_table}
\end{table*} 

\section{Experiments}
\subsection{Implementation details}

We use a publicly available open-source implementation of pointer-generator network
\footnote{https://github.com/atulkum/pointer\_summarizer}.

We implemented our LF2BERT model using the HuggingFace \cite{wolf2020transformers} library creating an encoder-decoder model with Longformer as an encoder and BERT as a decoder.

%For decoder we used the standard BERT configuration (12 layers, 12 heads, dimensions of all hidden states and word embeddings equal to 768). 
%The model was trained under softmax cross entropy loss with Adam optimizer and half precision.
%We used linear learning rate decay with 5e-5 initial learning rate. At generation we used beam size equal to 5. The models were trained using NVIDIA Tesla V100 GPUs.

We represent the patient history as a sequence of records starting with ['CLS'] token, records separated by ['SEP'] and ending with ['EOS']. Each record starts with a date and the title. To specify which one of three sections the model should generate, we add a section-speciﬁc (text) preﬁx to the input sequence of patient history before feeding it to the model. For example, to generate recommendations the input of the model would be the sequence “['CLS'] Recommendations: 01/24/2017 Initial examination: Local status: Walks without limp and additional support ..."
and would be trained to output "Observation of an orthopedic traumatologist, therapist. Dressings every other day. Remove stitches 10 days after surgery."

\subsection{Automated metrics evaluation}

We evaluate the quality of predictions using standard ROUGE metric \cite{lin2004rouge}, measuring the F1 scores for ROUGE-1, ROUGE-2 and ROUGE-L (word-overlap, bigram-overlap, and longest common sequence between evaluated summary and the target summary)

It is well-known that this metric has some limitations, primarily because it calculates the overlap of N-grams between the predicted and reference summaries and does not reflect the utility of the predictions. To mitigate this issue we conducted an evaluation with two board-certified doctors. 

\subsection{Doctor evaluation}

To evaluate the medical accuracy of our summarization model we conducted an experiment with two doctors assessing the model performance from the medical perspective. Each of the doctors was provided with 100 hospitalisation histories. The doctors were tasked to evaluate the ground truth text, the model predictions and a random prediction. For Treatment and Performed Labs sections, a random prediction is simply a subset of sentences from a patient's history. For Recommendations the random prediction is a real recommendation given to another patient.

For each hospitalisation history four texts where provided: 
\begin{itemize}
    \setlength\itemsep{-0.3em}
    \item ground truth as extracted from the summary section of the report
    \item a text selected randomly from a pool of all reports
    \item PGN model output
    \item LF2BERT model output.
\end{itemize}

The doctors were tasked with scoring all four summary variants on several criteria:
\begin{itemize}
    \setlength\itemsep{-0.3em}
    \item missing information (yes/no)
    \item incorrect information (yes/no)
    \item extra information (yes/no)
    \item grammatical and syntax correctness (yes/no)
    % \item overall quality score (on scale from 0 to 10)
\end{itemize}
Some of these criteria are of greater clinical relevance, and other criteria we did not measure. To mitigate this we asked doctors to give an overall quality score from 1 to 10 for each prediction, ground truth, and random text.

This task was repeated 3 times for each of the summary sections: treatment, performed labs, recommendations. 

\section{Results}

\subsection{Automated metrics}

The results of comparison between PGN and LF2BERT are presented in Table \ref{tab:valid_table}. We also included the scores for random summaries selected as described in the previous section. LF2BERT drastically improves on PGN for all summary sections and all ROUGE metrics. The improvement is about 20\% for the Treatment section, 80\%-110\% for the Performed labs section and 50\%-70\% for the Recommendations section. 

\begin{figure*}
\includegraphics[width=\textwidth]{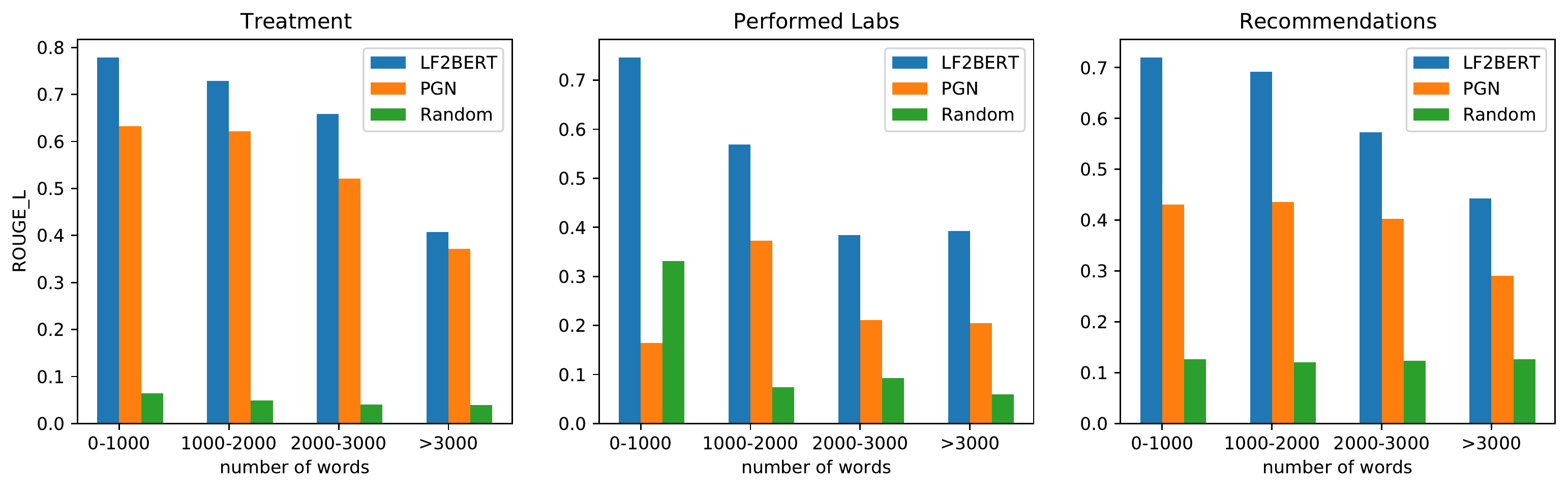}
\caption{Averaged ROUGE-L score with respect to different lengths of the patient's history.} 
\label{figure:rouge_length}
\end{figure*}

We also compared ROUGE-L scores for LF2BERT, PGN and random predictions with respect to length of patient's hospitalisation history. The results are shown in Figure \ref{figure:rouge_length}. LF2BERT performs better than PGN for all hospitalisation history lengths with more pronounced improvements for shorter hospitalisation histories.

\begin{table}[ht]
\centering
{\begin{tabular}{lccc}
\hline 
& Treatment & Labs & Rec. \\
\hline 
Human-written & 31 & 23 & 31 \\
\lfbert & 24 & 27 & 21 \\
Roughly equal & 45 & 50 & 48 \\
\hline 
% Human-written & 22 & 13 & 28 \\
% \lfbert & 25 & 30 & 20 \\
% Roughly equal & 51 & 57 & 52 \\
% \hline 
\end{tabular}}
\caption{Doctor evaluation of which text is better: the human-written or the \lfbert prediction. The scores are given for all sections: Treatment, Performed Labs and Recommendations. All scores are percentages.}
\label{tab:human_vs_machine}
\end{table} 

\subsection{Doctor evaluation}

\begin{table}[ht]
\centering
\small
{\begin{tabular}{l|cc|cc}
 & RND & Target & PGN & LF2BERT \\
\hline 
\multicolumn{5}{c}{Doctor 1} \\
\hline 
No missed info & 4.0 & 68.0 & \textbf{71.0} & 66.3\\
No wrong info & 2.0 & 38.0 & \textbf{56.0} & 41.8 \\
No extra info &23.0 & 53.0 & \textbf{73.0} & 59.2 \\
No text errors & 59.0 & 96.0 & 96.0 & \textbf{96.9}\\
Overall rating & 1.2 & 6.8 & \textbf{7.5} & 7.1\\
\hline
\multicolumn{5}{c}{Doctor 2} \\
\hline 
No missed info & 1.0 & 77.0 & 65.7 & \textbf{69.4}\\
No wrong info & 1.0 & 46.0 & \textbf{43.4} & 38.8 \\
No extra info & 80.2 & 91.0 & \textbf{90.9} & 89.8 \\
No text errors  & 91.7 & 94.0 & 90.9 & \textbf{93.9} \\
Overall rating & 1.1 & 6.2 & \textbf{5.7} & 5.5 \\

\end{tabular}}
\caption{Doctor evaluation of Treatment summary section} 
\label{tab:Table_treatment}
\end{table} 

\begin{table}[ht]
\centering
\small
{\begin{tabular}{l|cc|cc}
 & RND & Target & PGN & LF2BERT \\
\hline 
\multicolumn{5}{c}{Doctor 1} \\
\hline 
No missed info & 0.0 & 16.7 & \textbf{26.7} & 23.3 \\
No wrong info & 0.3 & 30.0 & \textbf{50.0} & 33.3 \\
No extra info & 0.3 & 60.0 & \textbf{73.3} &  \textbf{73.3}\\
No text errors & 83.3 & 96.7 & 93.3 & \textbf{100.0}\\
Overall rating & 1.2 & 4.1 & \textbf{5.6} & 5.0\\
\hline
\multicolumn{5}{c}{Doctor 2} \\
\hline 
No missed info & 3.3 & 70.0 & 56.7 & \textbf{70.0}\\
No wrong info & 3.3 & 56.7 & 46.7 & \textbf{46.7} \\
No extra info & 76.7 & 83.3 & 70.0 & \textbf{80.0} \\
No text errors  & 90.0 & 93.3 & 93.3 & \textbf{96.7}\\
Overall rating & 1.6 & 5.5 & 4.5 & \textbf{4.9} \\

\end{tabular}}
\caption{Doctor evaluation of Performed labs summary section} 
\label{tab:Table_labs}
\end{table}

\begin{table}[ht]
\centering
\small
{\begin{tabular}{l|cc|cc}
 & RND & Target & PGN & LF2BERT \\
\hline 
\multicolumn{5}{c}{Doctor 1} \\
\hline 
No missed info & 3.0 & 59.0 & 43.0 & \textbf{54.0} \\
No wrong info & 2.0 & 17.0 & \textbf{24.0} & 17.0\\
No extra info & 15.0 & 39.0 & \textbf{43.0} & 38.0 \\
No text errors & 89.0 & 96.0 & 96.0 & \textbf{97.0}\\
Overall rating & 1.4 & 5.8 & 4.7 & \textbf{5.7}\\
\hline
\multicolumn{5}{c}{Doctor 2} \\
\hline 
No missed info & 5.1 & 83.0 & 46.9 & \textbf{76.3}\\
No wrong info & 3.0 & 71.0 & 36.7 &  \textbf{56.7}\\
No extra info & 76.8 & 87.0 & \textbf{90.8} &  86.6\\
No text errors  & 97.0 & 96.0 & 95.9 & \textbf{97.9}\\
Overall rating & 1.4 & 7.4 & 4.9 & \textbf{6.5}\\

\end{tabular}}
\caption{Doctor evaluation of Recommendations summary section} 
\label{tab:Table_recommendations}
\end{table} 

The results are presented in Table \ref{tab:Table_treatment},  Table \ref{tab:Table_labs} and Table \ref{tab:Table_recommendations}

For all summmary sections LF2BERT performs consistently better on overall grammatical and syntax correctness beating not only PGN but also the original human-generated target.
For the treatment section PGN performs generally better.
For the labs section PGN and LF2BERT perform at comparable level with one doctor preferring the texts generated by PGN and another preferring the texts generated by LF2BERT.
For the recommendation section LF2BERT performs generally better.

One of the doctors scored both PGN and LF2BERT models outputs higher than the true human-generated target for treatments and labs sections. The same doctor scored LF2BERT as comparable to the true target for recommendation section (LF2BERT score of 5.7 vs target score of 5.8).

Note that the random prediction received a rate near 1 in every section, and the percentage of missing information and inaccurate/suspicious information is close to 100\%. We therefore believe the result suggests clinical validity of our experimental setting and testifies to the reliability of our experiment.

Additionally we use the overall rating to determine which of human-written and model-generated predictions is better. If the absolute rate difference between two predictions is less or equal to one, then they are considered "roughly equal", otherwise someone wins. The results are presented in Table \ref{tab:human_vs_machine}. According to this setup \lfbert predictions are at least as good as human-written texts in 69\%, 77\% and 69\% of cases in Treatment, Performed Labs and Recommendations sections accordingly. 

The results are presented in Table \ref{tab:Table_treatment},  Table \ref{tab:Table_labs} and Table \ref{tab:Table_recommendations}

\subsection{Case Study}

\begin{figure*}
\includegraphics[width=\textwidth]{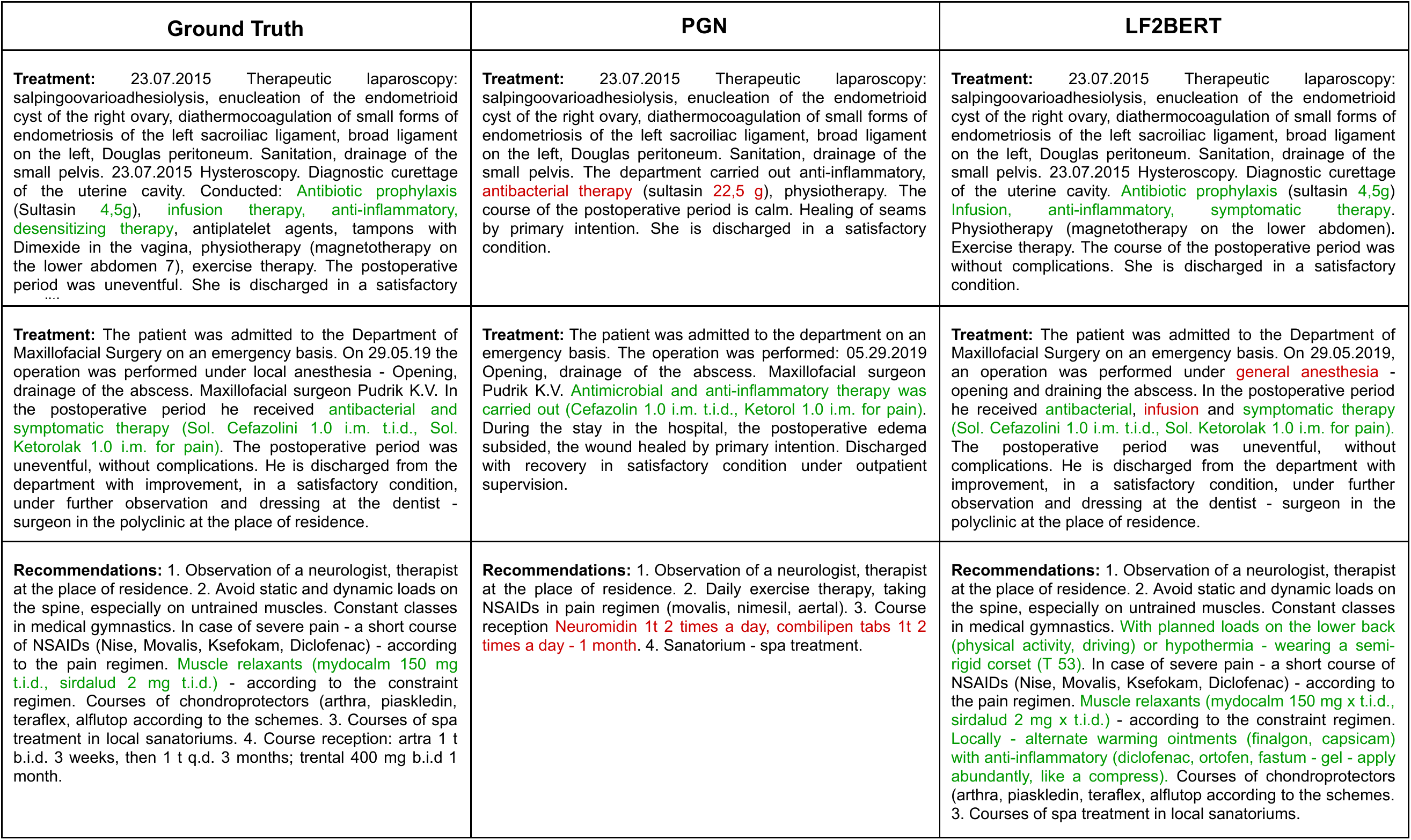}
\caption{Sampled test examples of sections generated by \pgn and \lfbert models compared with ground truth section. The first two rows are the examples of Treatment section, the third of Recommendations.} 
\label{figure:cases}
\end{figure*}

Along with the described quantitative experiments we perform qualitative analysis over three sample test cases (Figure \ref{figure:cases}). We compare the predictions of \pgn and \lfbert with the ground truth in two Treatment and one Recommendation sections. 

In the first case the summary of \lfbert completely coincides with the ground truth summary, despite the fact that different phrases were used. \pgn did not fully reflect the available information correctly - it mentioned "antibiotic therapy" instead of "antibiotic prophylaxis" and the dosage of the drug was also determined incorrectly. 

In the second case, both model generated Treatment section correctly and \pgn generated the summary slightly better. \pgn specified the therapy as "anti-inflammatory", instead of "symptomatic", which, taking into account the drugs, is not a mistake. \lfbert incorrectly specified the therapy as infusion. In addition, it incorrectly mentioned "general anesthesia" instead of "local anesthesia".

In the third case, the \pgn only superficially gave a set of recommendations, unlike \lfbert which gave recommendation in detail. Moreover, \pgn gave a different set of medicines, which is a serious mistake. \lfbert added some information that is not present in the ground truth, but in these cases it concretizes and does not distort what was stated in the ground truth. 

Due to space constraints, we could not include the complete patient history. However, we do show two examples in the original language with the complete patient history, Treatment section and the aggregated \lfbert attention visualization in appendix (Figure 5, Figure 6).

% Appendix (Figure \ref{figure:case1}, Figure \ref{figure:case2_1}).

\section{Conclusion}

In this work we introduced a novel model for neural abstractive medical summarization with an end-to-end transformer model. Our model is an encoder-decoder model with a Longformer model as an encoder and BERT as a decoder. The usage of the Longformer model allows our model to process long input sequences like hospitalisation histories.
We conducted experiments for three different summary sections (treatment, performed labs, recommendations) using both automated evaluation metrics (ROUGE-1, ROUGE-2, ROUGE-L) and certified doctor evaluation.

Our experiments show significant improvements on the medical summarization task over the current state-of-the-art approach – pointer-generator networks when measured on ROUGE metrics for all summary sections and all input lengths.
Doctor evaluation showed that our model not only shows consistent improvements over the PGN network on overall grammatical and syntax correctness of generated texts, but is actually rated higher on this criterion than the original targets for all three sections. When compared on overall medical quality of summaries our model clearly outperforms the baseline for the recommendations section while being comparable for the treatment section.


%\bibliographystyle{acl_natbib}

%\appendix

%\section{Appendix}
%\label{sec:appendix}

%\begin{figure*}
%\includegraphics[width=\textwidth]{images/case1.png}
%\caption{Example 1. The patient history with the %aggregated \lfbert attention and the corresponding %Treatment section. The TARGET denotes the Ground Truth %treatment section, PREDICTED denotes \lfbert %prediction.}
%\label{figure:case1}
%\end{figure*}

%\begin{figure*}
%\includegraphics[width=\textwidth]{images/case2_1.png}
%\caption{Example 2, first part.}
%\label{figure:case2_1}
%\end{figure*}

%\begin{figure*}
%\includegraphics[width=\textwidth]{images/case2_2.png}
%\caption{Example 2, second part.}
%\label{figure:case1_2}
%\end{figure*}

\end{document}